\pdfoutput=1
\documentclass[11pt]{article}

\usepackage[preprint]{acl}

\usepackage{times}
\usepackage{latexsym}

\usepackage[T1]{fontenc}

\usepackage[utf8]{inputenc}

\usepackage{microtype}

\usepackage{inconsolata}

\usepackage{graphicx}
\usepackage{amsmath}
\usepackage{subcaption}
\usepackage{booktabs,multirow}
\usepackage{adjustbox}
\usepackage[table]{xcolor} 
\definecolor{pedbg}{RGB}{235,245,255}

\usepackage{amsfonts}
\usepackage{listings}
\usepackage{comment}
\usepackage{xurl}

\newcommand{\stitle}[1]{\vspace{.8ex}\noindent{\textbf{#1}}}

\lstdefinestyle{promptstyle}{
    basicstyle=\ttfamily\footnotesize, 
    breaklines=true,                   
    breakatwhitespace=true,            
    frame=single,                      
    rulecolor=\color{black!20},        
    backgroundcolor=\color{gray!5},    
    columns=fullflexible,              
    keepspaces=true,
    showstringspaces=false,
    xleftmargin=3pt,                   
    xrightmargin=3pt,
    aboveskip=1em,
    belowskip=1em
}

\title{Parallel Context-of-Experts Decoding for Retrieval Augmented Generation}

\author{
Giulio Corallo
  \\
  SAP Labs, France
  \\
  EURECOM, France
  \\
  \texttt{giulio.corallo@sap.com}
  \And
Paolo Papotti
  \\
EURECOM, France
  \\
  \texttt{papotti@eurecom.fr}
}

\begin{document}
\maketitle
\begin{abstract}
Retrieval Augmented Generation faces a trade-off: concatenating documents in a long prompt enables multi-document reasoning but creates prefill bottlenecks, while encoding document KV caches separately offers speed but breaks cross-document interaction. We propose \textit{Parallel Context-of-Experts Decoding} (\textsc{Pced}), a training-free framework that shifts evidence aggregation from the attention mechanism to the decoding. \textsc{Pced} treats retrieved documents as isolated "experts", 
synchronizing their predictions via a novel retrieval-aware contrastive decoding rule that weighs expert logits against the model prior. This approach recovers cross-document reasoning capabilities without constructing a shared attention across documents.
\end{abstract}

\section{Introduction}
Retrieval Augmented Generation (RAG) augments language models with external corpora to improve factuality and reduce hallucinations~\citep{NEURIPS2020_6b493230, gao2023retrieval, 10.1145/3637528.3671470}. 
However, standard pipelines concatenate many retrieved documents into a single \textit{long context} prompt, making inference dominated by prefill latency~\citep{10.1145/3600006.3613165, 10.5555/3691938.3691949, cheng2025lmcache}. 
Additionally, long contexts increase reasoning failures, as models often struggle to integrate evidence spread across multiple documents~\citep{liu-etal-2024-lost}.

\begin{figure}[th]
  \centering
\includegraphics[width=\columnwidth]{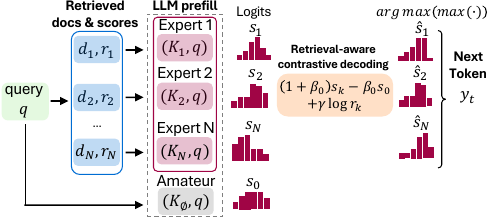}
  \caption{
  Parallel Context-of-Experts Decoding (\textsc{Pced})
  runs one expert per retrieved document (and a no-context, amateur prior) in parallel and chooses each next token based on retrieval support, enabling evidence to be stitched across documents without joint attention.} 
  \label{fig:pced}
  \vspace{-2ex}
\end{figure}

Parallel KV cache encoding mitigates prefill cost by encoding retrieved documents independently and reusing their cached states at inference time~\citep{yang2025kvlink,yang2025ape}. 
However, removing cross-document attention during encoding can substantially degrade performance on multi-hop and reasoning-intensive queries~\citep{10.1145/3689031.3696098}.

We propose \textbf{Parallel Context-of-Experts Decoding} (\textsc{Pced}), a training-free framework that shifts document aggregation from attention to decoding. 
As depicted in Figure~\ref{fig:pced}, at each generation step, \textsc{Pced} treats each document as a separate “expert”, which proposes a next-token distribution from its own KV cache, and then weights the best-supported token so evidence can be \textit{efficiently} aggregated across documents 
without building a joint attention context.
We make three contributions: (1) a \textit{parallel, modular KV cache framework} with decode-time evidence aggregation; (2) \textit{token-level expert switching} to recover cross-document reasoning via dynamic expert selection \emph{at every token step} without shared attention; and (3) \textit{retrieval-integrated priors} that inject scalar scores into the contrastive decoding to gate noise from irrelevant experts.
On benchmarks like LOFT and LongBench, \textsc{Pced} outperforms prior parallel methods by up to 70 points and often matches or outperforms \textit{long context} baselines, while delivering over $180\times$ speedup in time-to-first-token.

\section{Related Work}
We position our work at the intersection of (1) KV caching for parallel prefill, (2) cross-document interaction recovery under independent KV caches, and (3) context-aware decoding.

\stitle{Parallel encoding} eliminates prefill cost by precomputing \emph{offline} per-document KV caches that can be retrieved at inference time.
Prior work includes training-free masking for blockwise/parallel attention~\citep{ratner-etal-2023-parallel}, fine-tuning to mitigate quality degradation under blocked attention~\citep{ma2025blockattention}, and interfaces that decouple document encoding from generation~\citep{yen-etal-2024-long}.
Systems work integrates KV cache retrieval into RAG pipelines~\citep{lu-etal-2025-turborag}.
These approaches assume documents as independently encodable, while 
we study how to aggregate evidence across multiple cached documents at inference.

\stitle{Cache merging} techniques
encode documents independently and then aim to restore the cross-document attention, 
as simply concatenating per-document KV caches does not recover it~\citep{10.1145/3689031.3696098}.
Recent methods achieve this via selective recomputation at merging~\citep{10.1145/3689031.3696098}, learned bridging tokens for 
inter-document interactions~\citep{yang2025kvlink}, or training-free alignment to approximate sequential attention (APE)~\citep{yang2025ape}.
Our work 
preserves per-document modularity while enabling effective cross-document reasoning.

\stitle{Context-aware decoding (CAD)}~\citep{shi-etal-2024-trusting} improves faithfulness by shifting probability mass toward tokens supported by context; it is related to contrastive decoding~\citep{li-etal-2023-contrastive} and classifier-free guidance in diffusion models~\citep{ho2021classifierfree}.
However, most CAD formulations assume a single supportive context that defines the conditional distribution.
DvD~\citep{jin-etal-2024-dvd} extends CAD to multiple documents but collapses them into a single input sequence, which conflicts with per-document KV cache reuse, where documents must be encoded separately.

\section{Methodology}
We introduce {Parallel Context-of-Experts Decoding (\textsc{Pced})}, a training-free framework for scalable and faithful multi-document generation.
RAG pipelines typically employ a two-stage process: retrieving candidate documents using 
embeddings to maximize \emph{recall}, followed by a cross-encoder reranker to reorder candidates and maximize \emph{precision}. Crucially, the scalar relevance scores produced during these stages are used only for document selection and then discarded. We argue that this discards valuable evidence about how strongly each document should be trusted during decoding.
\textsc{Pced} converts these scores into \textit{a document-level prior that controls how much each expert influences the next-token distribution}, via a novel retrieval-aware contrastive decoding criterion.

\stitle{Offline KV cache preparation.}
Following prior cache-augmented generation work~\citep{10.1145/3701716.3715490, lu-etal-2025-turborag, yang2025ape, 10.1145/3768628}, we assume a datastore $\mathcal{DB}$ over a corpus $\mathcal{D}$ that stores, for each document $d_i$, an embedding $\mathbf{e}_i$ for retrieval and its precomputed KV cache $\mathbf{K}_i$:
\begin{equation}
\mathcal{DB}=\{(d_i,\mathbf{e}_i,\mathbf{K}_i)\}_{i=1}^{|\mathcal{D}|}.
\end{equation}

\stitle{Retrieval and relevance scoring.}
Given a query $q$, we retrieve the top-$N$ documents and obtain retrieval scores $\mathbf{r}^{\text{ret}}=(r^{\text{ret}}_1,\ldots,r^{\text{ret}}_N)$. We then rerank these documents with a cross-encoder, producing reranker scores $\mathbf{r}^{\text{rer}}=(r^{\text{rer}}_1,\ldots,r^{\text{rer}}_N)$.
We map both score sets to the range $[0,1)$. Since $r^{\text{ret}}$ primarily reflects recall and $r^{\text{rer}}$ precision, we fuse them into a single per-document relevance score via the harmonic mean 
$r_k=\frac{2\,r^{\text{ret}}_k\,r^{\text{rer}}_k}{r^{\text{ret}}_k+r^{\text{rer}}_k},
k\in\{1,\ldots,N\}$.

\stitle{Parallel Context-of-Experts.}
As depicted in Figure~\ref{fig:pced}, \textsc{Pced} operates on $N{+}1$ parallel streams (\emph{experts}) in a single batched forward pass:
one \textbf{amateur} expert with an empty cache $\mathbf{K}_0=\emptyset$ (model prior)
and $N$ \textbf{contextual} experts, one per retrieved document, with caches $\mathbf{K}_{1:N}$ and associated relevance scores $r_{1:N}$.
Given batch $\mathcal{B}=\{\mathbf{K}_k\}_{k=0}^{N}$, processing the query $q$ updates all experts' caches in parallel. At each step, this yields per-expert logits
$s_k \in \mathbb{R}^{|\mathcal{V}|}$ over the vocabulary $\mathcal{V}$.

\paragraph{Retrieval-aware contrastive decoding.}
For each contextual expert $k\in\{1,\ldots,N\}$, we calibrate logits against the amateur $s_0$ and incorporate a retrieval-based prior:
\begin{equation} \label{eq:contrastive-retrieval-bias-wrapped} \hat{s}_k = \underbrace{(1+\beta_0)\,s_k-\beta_0\,s_0}_{\substack{\text{Contrastive}\\\text{decoding}}} \;+\; \underbrace{\gamma\,\log r_k}_{\substack{\text{Retrieval}\\\text{prior}}} \end{equation}
Here, $\beta_0$ controls contrast strength between amateur and expert, and $\gamma$ controls retrieval gating. We compute $\beta_0$ dynamically as in AdaCAD~\citep{wang-etal-2025-adacad} for the \emph{first} generated token 
and keep it fixed thereafter. We empirically set $\gamma=2.5$ for all experiments (ablations in Appendix~\ref{sec:beta_ablation} for $\beta$, \ref{sec:gamma_ablation} for $\gamma$).
Finally, the next token $y_t$ is the one with the highest score among all experts’ candidates.
\begin{equation}
    y_t = \arg\max_{v \in \mathcal{V}} \left( \max_{k \in \{1,\dots,N\}} \hat{s}_k(v) \right)
\end{equation}
The chosen token is appended to the \textbf{shared generation history} for all experts at each step.

\begin{table*}[t]
  \centering
  \footnotesize
  \setlength{\tabcolsep}{2pt}
  \renewcommand{\arraystretch}{1.15}
  \caption{\textbf{Main results on RAG and ICL benchmarks.} We compare our Parallel Expert Decoding (\textsc{Pced}) framework, equipped with Sparse, Dense, or ColBERT experts, against KV merging (APE), agentic (MapReduce), and standard concatenation baselines. \emph{Corpus in Ctx (All)} is the baseline with all retrieved candidates in context.} 
  \label{tab:main_results_sidebyside}

  \begin{subtable}[t]{0.48\textwidth}
    \centering
    \caption{\textsc{Mistral-Nemo-13B-Instruct}}
    \begin{adjustbox}{max width=\linewidth}
      \begin{tabular}{@{}l l
        c 
        c 
        >{\columncolor{pedbg}}c 
        >{\columncolor{pedbg}}c 
        >{\columncolor{pedbg}}c 
        c 
        c 
      @{}}
        \toprule
        & & \multicolumn{1}{c}{KV Merge} & \multicolumn{1}{c}{Agentic}
          & \multicolumn{3}{c}{\cellcolor{pedbg}PCED} & \multicolumn{2}{c}{Corpus in Ctx} \\
        \cmidrule(lr){3-3}\cmidrule(lr){4-4}\cmidrule(lr){5-7}\cmidrule(lr){8-9}
        Task & Dataset & {APE} & {MapRed.} & {\cellcolor{pedbg}Sparse} & {\cellcolor{pedbg}Dense} & {\cellcolor{pedbg}ColBERT} & {Single} & {All} \\
        \midrule
        \multirow{5}{*}{\textbf{RAG}}
          & \textsc{HotpotQA} & 27.0 & 56.0 & 65.0 & \textbf{66.0} & 66.0 & 54.0 & 64.0 \\
          & \textsc{MuSiQue}  & 11.0 & 26.0 & \textbf{36.0} & 34.0 & 35.0 & 17.0 & 28.0 \\
          & \textsc{NQ}       & 38.0 & 62.0 & 80.0 & \textbf{81.0} & 81.0 & 60.0 & 76.0 \\
          & \textsc{QAMParI}  & 7.0 & \textbf{85.0} & 75.0 & 71.0 & 71.0 & 75.0 & 74.0 \\
              & \textsc{Quest}    & 1.0 & 42.0 & \textbf{55.0} & 54.0 & 54.0 & 38.0 & 19.0 \\
        \addlinespace[2pt]
        \midrule
        \multirow{4}{*}{\textbf{ICL}}
          & Web          & 58.9 & 42.2 & 61.1 & \textbf{62.2} & 62.2 & 35.6 &  61.1 \\
          & Tracking7    & 6.7 & \textbf{13.3} & 7.8 & 7.8 & 7.8 & 10.0 & 6.7 \\
          & Date         & 40.0 & 55.6 & 57.8 & \textbf{57.8} & 57.8 & 57.8 & 54.4 \\
        \bottomrule
      \end{tabular}
    \end{adjustbox}
  \end{subtable}
  \hfill
  \begin{subtable}[t]{0.48\textwidth}
    \centering
    \caption{\textsc{Llama-3.1-8B-Instruct}}
    \begin{adjustbox}{max width=\linewidth}
      \begin{tabular}{@{}l l
        c 
        c 
        >{\columncolor{pedbg}}c 
        >{\columncolor{pedbg}}c 
        >{\columncolor{pedbg}}c 
        c 
        c 
      @{}}
        \toprule
        & & \multicolumn{1}{c}{KV Merge} & \multicolumn{1}{c}{Agentic}
          & \multicolumn{3}{c}{\cellcolor{pedbg}PCED} & \multicolumn{2}{c}{Corpus in Ctx} \\
        \cmidrule(lr){3-3}\cmidrule(lr){4-4}\cmidrule(lr){5-7}\cmidrule(lr){8-9}
        Task & Dataset & {APE} & {MapRed.} & {\cellcolor{pedbg}Sparse} & {\cellcolor{pedbg}Dense} & {\cellcolor{pedbg}ColBERT} & {Single} & {All} \\
        \midrule
        \multirow{5}{*}{\textbf{RAG}}
          & \textsc{HotpotQA} & 16.0 & 41.0 & 64.0 & \textbf{64.0} & 64.0 & 49.0 &  66.0 \\
          & \textsc{MuSiQue}  & 4.0  & 8.0  & 14.0 & \textbf{21.0} & 21.0 & 7.0  & 16.0 \\
          & \textsc{NQ}       & 9.0  & 50.0 & 83.0 & \textbf{85.0} & 85.0 & 58.0 & 79.0 \\
          & \textsc{QAMParI}  & 7.0 & 68.0 & \textbf{77.0} & 76.0 & 76.0 & 72.0 & 86.0 \\
          & \textsc{Quest}    & 0.0 & 41.0 & \textbf{45.0} & 40.0 & 40.0 & 39.0 & 44.0 \\
        \addlinespace[2pt]
        \midrule
        \multirow{4}{*}{\textbf{ICL}}
          & Web          & 61.1 & 56.7 & 62.2 &  \textbf{64.4} & 63.3 & 57.8 & 57.8 \\
          & Tracking7    & 3.3 & \textbf{13.3} & 11.1 & 11.1 & 11.1 & 11.1 & 7.8 \\
          & Date         & 0.0 & 44.4 & \textbf{53.3} & 47.8 & 48.9 & 51.1 &  53.3 \\
        \bottomrule
      \end{tabular}
    \end{adjustbox}
  \end{subtable}
\end{table*}

\section{Experimental Setup}
We test \textsc{Pced} on RAG, In Context Learning (ICL), and long-context QA with distractors. For all methods, we fix the LLM, prompts, and retrieved candidates; varying only how context is incorporated. 

\stitle{Datasets and Metrics.}
We use the LOFT benchmark~\cite{Lee2024LongContext} for RAG and ICL. We retrieve a fixed pool of the top-90 documents per query, shared across all baselines. Performance is measured via \textit{Subspan Exact Match} for RAG tasks and \textit{Exact Match} for ICL tasks.
We also evaluate on the query-focused LongBench subsets~\cite{bai-etal-2024-longbench} using official metrics. To test robustness to irrelevant context, we concatenate the \textit{gold} document with $K{=}2$ uniformly sampled \textit{distractors} from other test samples, keeping the corpus-in-context baseline under 128k tokens.

\stitle{LLMs.}
We report main results with \textsc{Mistral-Nemo-13B-Instruct}~\citep{mistral-nemo-2024} and \textsc{Llama-3.1-8B-Instruct}~\citep{grattafiori2024llama}, and LongBench results with \textsc{Qwen3-8B}~\citep{yang2025qwen3} extended to 128k tokens with \textsc{YaRN}~\citep{peng2024yarn}. Decoding is greedy for all methods.

\stitle{\textsc{Pced} variants.}
We evaluate three scoring variants: \textit{Sparse}, \textit{Dense}, and \textit{ColBERT}. The set of retrieved documents is {identical} for all methods. These variants differ only in the relevance signal $r_k$ extracted from \texttt{bge-m3} to weight experts in Eq.~\ref{eq:contrastive-retrieval-bias-wrapped}.

\stitle{Baselines.}
We compare against three baseline families. {Standard concatenation} (\emph{Corpus in Ctx}) conditions on either the \textbf{Single} top-1 document retrieved or \textbf{All} retrieved documents in a single prompt (e.g., top-$90$ for LOFT). \textit{KV cache merging} (\textsc{\textbf{APE}}), prefills each document independently and merges the resulting KV caches. 
\textit{Agentic aggregation} (\textsc{\textbf{MapReduce}}) performs per-document summarization (map) followed by a final QA aggregation step (reduce)~\citep{zhou-etal-2025-llmxmapreduce}.

\begin{table*}[t]
    \centering
    \footnotesize
    \caption{\textbf{Results on LongBench using \textsc{Qwen3-8B}.} \textsc{Pced} against the full-context baseline Corpus in Ctx (All).}
    \label{tab:longbench_results_qwen}
    \begin{adjustbox}{max width=0.9\textwidth}
    \begin{tabular}{l ccc ccc c c c}
        \toprule
        & \multicolumn{3}{c}{\textbf{Single-Doc QA}} 
        & \multicolumn{3}{c}{\textbf{Multi-Doc QA}} 
        & \textbf{Summ.} 
        & \textbf{Few-Shot} 
        & \textbf{Code} \\
        \cmidrule(lr){2-4} \cmidrule(lr){5-7} \cmidrule(lr){8-8} \cmidrule(lr){9-9} \cmidrule(lr){10-10}
        \textbf{Method} & 
        \textsc{NarQA} & \textsc{Qasper} & \textsc{MultiF} & 
        \textsc{Hotpot} & \textsc{2Wiki} & \textsc{MuSiQue} & 
        \textsc{QMSum} & \textsc{TriviaQA} & \textsc{RepoB-P} \\
        \midrule
        
        Corpus in Ctx (All) & 
        21.1 & 25.2 & 52.8 & 
        56.3 & 44.2 & 25.3 & 
        22.0 & 
        84.0 & 
        51.1 \\ 

        \rowcolor{pedbg}
        PCED (Sparse) & 
        25.1 & 24.2 & \textbf{53.0} & 
        62.1 & 49.4 & \textbf{33.4} & 
        22.7 & \textbf{88.8} & 59.7 \\
        
        \rowcolor{pedbg}
        PCED (Dense) & 
        \textbf{25.4} & \textbf{25.7} & 52.6 & 
        \textbf{62.6} & \textbf{49.4} & 33.3 & 
        \textbf{22.9} & 88.2 & \textbf{60.1} \\

        \rowcolor{pedbg}
        PCED (ColBERT) & 
        25.4 & 25.7 & 52.6 & 
        62.6 & 49.4 & 33.3 & 
        22.9 & 88.2 & 60.1 \\
        \bottomrule
    \end{tabular}
    \end{adjustbox}
        \vspace{-2ex}
\end{table*}

\section{Results and Discussion}
\label{sec:results}
We analyze our results around three main themes: (1) multi-document RAG and ICL with many candidate documents/exemplars, (2) single-document with query-focused understanding and generation tasks (including QA, summarization, code completion, and few-shot inference), and (3) efficiency.

\stitle{Cross-Document Reasoning Emerges at Decode Time.} In Table~\ref{tab:main_results_sidebyside},
\textsc{Pced} consistently outperforms KV cache merging (\textsc{APE}) in 
QA benchmarks that require aggregating evidence across multiple 
documents (e.g., \textsc{HotpotQA}, \textsc{MuSiQue}, \textsc{QAMParI}, \textsc{Quest}), and in ICL settings where exemplars must be used jointly. 
For instance, on \textsc{Llama-3.1-8B} \textsc{QAMParI}, \textsc{Pced} improves from 7 (\textsc{APE}) to {77} (\textsc{Pced-Sparse}), and yields up to {+23} points over \textsc{MapReduce} (e.g., \textsc{HotpotQA}). 
Moreover, \textsc{Pced} variants often match or exceed full-context concatenation: \textsc{Pced-Dense} outperforms Corpus in Ctx (All) in \textbf{11/16} settings despite encoding each document independently. These results suggest that much of the benefit of cross-document interaction can be recovered at \emph{decode time}.

\begin{figure}[h]
    \centering
    \includegraphics[width=0.99\columnwidth]{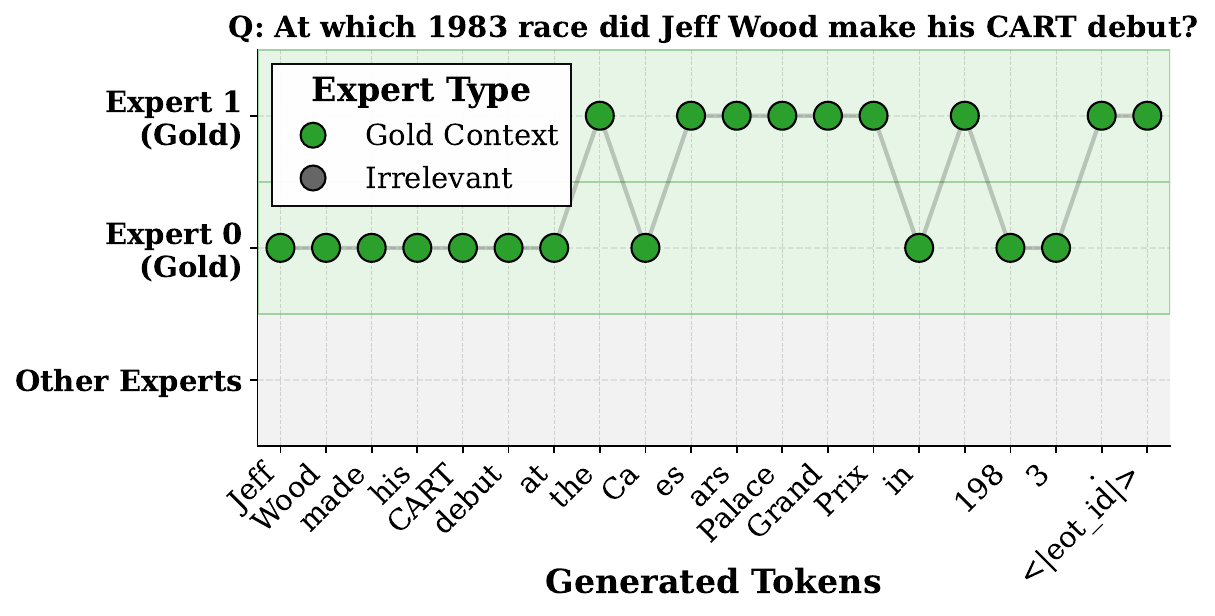}
    \caption{HotpotQA expert trace. Green dots illustrate the model hopping between multiple gold documents.}
    \label{fig:trace_hotpot}
\end{figure}

Figure~\ref{fig:trace_hotpot} illustrates this mechanism: to resolve a multi-hop query, the model first locks onto an expert containing the bridging entity, then pivots to a second expert for the final answer. By appending the chosen token to \emph{all} experts' shared generation histories, \textsc{Pced} effectively stitches evidence across isolated documents without shared attention.
We note that \textsc{MapReduce} remains superior on some settings (e.g., \textsc{QAMParI} with Mistral), suggesting cases where global synthesis across many documents is beneficial. However, \textsc{MapReduce} relies on multiple LLM calls (per-document summarization and an aggregation pass), while \textsc{Pced} aggregates evidence within a single decoding procedure.

\stitle{Less Noise, More Accuracy.}
\textsc{Pced} also improves performance on tasks where the answer is primarily supported by a \emph{single} document, but must be recovered from a large candidate pool. In these settings, full-context concatenation can degrade because relevant evidence is diluted by many near-miss documents and distractors, making attention noisier. By contrast, \textsc{Pced} isolates evidence by treating each document as an independent expert and explicitly emphasizing per-document relevance via retrieval-aware contrastive decoding (Eq.~\ref{eq:contrastive-retrieval-bias-wrapped}), which downweights irrelevant experts. 
Table~\ref{tab:main_results_sidebyside} shows that this yields strong gains on \textsc{NQ} under LOFT: with {Llama}, \textsc{Pced-Dense} improves from 58 (Corpus in Ctx Single) and 79 (All) to {85}, similarly with \textsc{Mistral} from 60 (Single) and 76 (All) to {81}. 
We observe the same trend in LongBench (Table~\ref{tab:longbench_results_qwen}): when the gold evidence is surrounded by irrelevant context, \textsc{Pced} benefits from expert isolation.

\begin{figure}
    \centering
    \includegraphics[width=\linewidth]{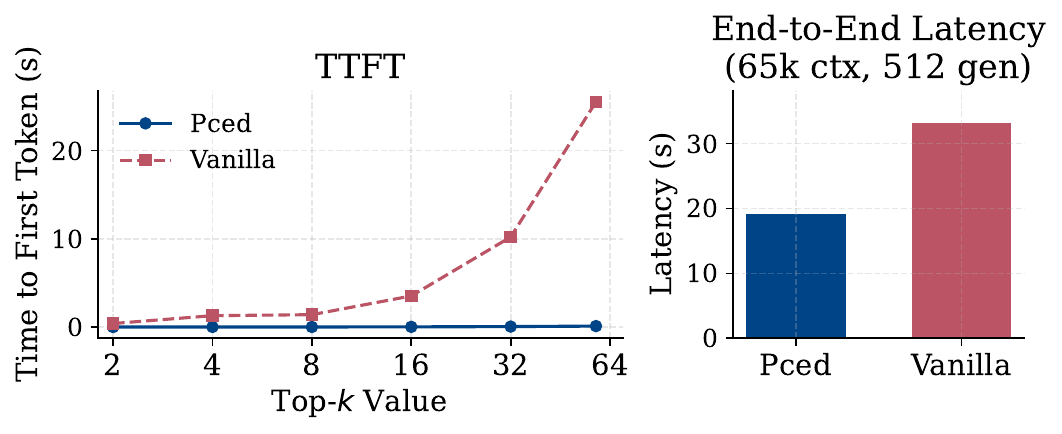}
    \caption{\textbf{Latency Benchmarks.} Comparison of TTFT scalability across Top-$k$ values (left) and total end-to-end latency with 65k context (right).}
    \label{fig:ttft}
    \vspace{-2ex}
\end{figure}

\stitle{Efficiency at Scale.}
Unlike context concatenation, which incurs high prefill costs, \textsc{Pced} leverages offline, reusable KV caches to reduce Time-To-First-Token (TTFT). As shown in Figure~\ref{fig:ttft}, \textsc{Pced} consistently achieves substantially lower TTFT across all top-$K$, with gains that scale to over \textbf{$180\times$} faster TTFT (0.14s vs.\ 25.50s). On long-context workloads (65k context tokens, 512 generated tokens), it yields a $\sim$1.7$\times$ reduction in end-to-end latency.
All results use a high-throughput setup with continuous batching and PagedAttention~\citep{10.1145/3600006.3613165} for both methods, validating the method's efficiency under realistic conditions.
\begin{table}[h]
    \centering
    \caption{\textbf{Component Analysis.} Disentangling benefits of Contrastive Decoding vs. Retrieval Prior.}
    \label{tab:ablation_components_main}
    
    \resizebox{\columnwidth}{!}{
    \setlength{\tabcolsep}{2pt}
    \begin{tabular}{ll ccc}
        \toprule
        & & \textbf{Only Contrastive} & \textbf{Only Retrieval} & \textbf{PCED} \\
                & & \textbf($\gamma=0$) &  ($\beta=0$) & \\
        \midrule
        \multirow{2}{*}{\textsc{Llama-8B}} 
          & \textsc{HotpotQA} & 46 & 53 & \textbf{64} \\
          & \textsc{NQ}       & 52 & 70 & \textbf{85} \\
        \addlinespace[0.3em] 
        \multirow{2}{*}{\textsc{Mistral-13B}} 
          & \textsc{HotpotQA} & 57 & 65 & \textbf{66} \\
          & \textsc{NQ}       & 71 & 80 & \textbf{81} \\
        \bottomrule
    \end{tabular}
    }
\end{table}

\stitle{Ablations.}
We verify that both terms in Eq.~\ref{eq:contrastive-retrieval-bias-wrapped} are important: removing the retrieval prior ($\gamma{=}0$) or the contrastive calibration ($\beta{=}0$) leads to large accuracy drops (Table~\ref{tab:ablation_components_main}).
We further find that Max aggregation best supports token-level expert switching in multi-hop QA, whereas soft mixtures can help in single-doc settings.
Full sweeps over $\beta,\gamma$, aggregation rules, and top-$k$ are in Appendix~\S\ref{sec:ablations}.


\section{Conclusion}
We presented \textsc{Pced}, a training-free decoding framework that enables efficient, multi-document reasoning under parallel, cache-native conditioning. \textsc{Pced} replaces long-context attention with retrieval-aware expert logit fusion at decode time, preserving KV cache modularity while recovering cross-document reasoning. Empirically, it matches or surpasses full-context baselines and is more robust to distractors. This offers an exciting alternative to long context models, allowing the number of documents to scale flexibly with batch size rather than being limited by the training context window.
\section*{Limitations}
Despite its strong empirical performance and efficiency benefits, \textsc{Pced} has several limitations.

\paragraph{Dependence on access to model logits.}
\textsc{Pced} relies on per-expert token-level logits to perform retrieval-aware contrastive decoding, explicitly calibrating contextual experts against the amateur (prior) expert at each decoding step. This requirement assumes full access to the model’s output logits. As a result, \textsc{Pced} cannot be directly applied to closed-source or API-only language models that expose only sampled tokens or log-probabilities for a limited subset of candidates. While this constraint is shared by many contrastive and guidance-based decoding methods, it currently restricts the applicability of \textsc{Pced} to open or self-hosted models.

\paragraph{Sensitivity to retrieval quality.}
Like most RAG approaches, \textsc{Pced} depends on the quality of the retrieved documents and their associated relevance scores. If relevant evidence is not retrieved or is assigned low relevance, the corresponding expert may be underweighted or never selected during decoding. Although retrieval-aware contrastive decoding mitigates noise from weak or irrelevant documents, it cannot recover evidence that is entirely absent from the candidate set. That said, our formulation highlights an interesting direction for future work: rather than relying on external retrieval and reranking models, one could explicitly train language models to accept parallel contextual inputs and to learn, at each next token, which input to attend to. Such an approach could reduce reliance on external retrieval pipelines and enable end-to-end learning of expert selection and aggregation, enabling parallelization at inference.

\paragraph{Storage-Computation Trade-offs.}
\textsc{Pced} accelerates inference by effectively offloading online computation to offline storage. By persisting precomputed KV caches, the framework eliminates runtime encoding latency; however, this imposes a storage footprint that scales linearly with both corpus size and hidden state dimensionality. For instance, storing FP16 KV caches for the LOFT \textsc{HotpotQA} corpus (1,222 passages of 74 tokens on average) using \textsc{Llama-3.1-8B} necessitates approximately 11.04 GB of storage. Consequently, \textsc{Pced} is optimally deployed in \textbf{read-heavy, write-rare} settings involving static corpora—such as enterprise knowledge bases—where the amortized storage cost is justified by the substantial reduction in query-time latency.

\bibliography{custom}

\appendix

\section{Evaluation Setup}
\label{sec:evaluation}

This appendix details the prompt templates and instantiation protocols for each dataset. To ensure a fair comparison \textbf{across all methods} (Concatenation, KV-merge, MapReduce, and \textsc{Pced}), we fix the underlying dataset fields, system prompt, context template, question template, and answer prefix, and vary \emph{only} the mechanism of context incorporation.
All experiments were executed with a fixed random seed (42) to ensure deterministic results. Unless otherwise stated, all reported numbers correspond to a single deterministic run per method.
\paragraph{Prompt Definitions.}
Each dataset instance is composed of four standardized fields: a \texttt{system\_prompt} containing high-level instructions; a \texttt{context\_template} which wraps the retrieved text; and a \texttt{question\_template} applied to the user query.

\subsection{LOFT Benchmark}
\label{sec:appendix-loft}
We utilize the LOFT benchmark~\citep{Lee2024LongContext}. 
Dataset statistics (e.g., number of examples, context lengths, and task distributions) are reported in Table~1 of the original LOFT paper~\citep{Lee2024LongContext}.

\subsubsection{LOFT-RAG Templates}
\label{sec:appendix-loft-rag}
For RAG tasks, all methods utilize the prompt configuration defined in Figure~\ref{fig:loft_rag_prompt}. The \texttt{\{context\}} slot is populated according to the specific method.

\begin{figure}[h]
\centering
\begin{footnotesize}
\begin{tabular}{|p{0.95\linewidth}|}
\hline
\textbf{System Prompt} \\
You will be given a list of documents. You need to read carefully and understand all of them. Then you will be given a query, and your goal is to answer the query based on the documents you have read. \\
\hline
\textbf{Context Template} \\
\texttt{\{context\}} \\
\hline
\textbf{Question Template} \\
Based on the documents above, can you answer the following query? Write a concise answer.\\
query: \texttt{\{question\}} \\
\hline
\end{tabular}
\end{footnotesize}
\caption{Prompt template configuration for LOFT-RAG tasks.}
\label{fig:loft_rag_prompt}
\end{figure}

\subsubsection{LOFT-ICL Templates}
\label{sec:appendix-loft-icl}
For In-Context Learning (ICL) tasks, we enforce a strict output format to facilitate automated parsing. The templates are defined in Figure~\ref{fig:loft_icl_prompt}.

\begin{figure}[h]
\centering
\begin{footnotesize}
\begin{tabular}{|p{0.95\linewidth}|}
\hline
\textbf{System Prompt} \\
Please answer the following questions and ensure you follow a consistent format. In particular, ensure your final answer always looks like `Output: ['your\_answer\_here']` After Output write ONLY the best option following the example(s). Do NOT write anything else. \\
\hline
\textbf{Context Template} \\
Example(s):\\
\texttt{\{context\}} \\
\hline
\textbf{Question Template} \\
Now begin!\\
\texttt{\{question\}} \\
\hline
\end{tabular}
\end{footnotesize}
\caption{Prompt template configuration for LOFT-ICL tasks.}
\label{fig:loft_icl_prompt}
\end{figure}

\subsection{Method-Specific Instantiations}

\paragraph{\textsc{Pced}.}
\textsc{Pced} treats retrieved documents (RAG) or exemplars (ICL) as independent contextual experts.
Concretely, for each query we create $N$ contextual expert inputs by applying the dataset \texttt{system\_prompt} and \texttt{context\_template} to documents, yielding $N$ separate (system, context) prompt instances.
At decoding time, each expert produces logits conditioned on its own KV cache.
We additionally include an \textbf{amateur} expert that represents the model prior: it is instantiated using \texttt{system\_prompt} only. All experts share the identical \texttt{question\_template}.

\paragraph{MapReduce.}
This method involves a two-stage process. First, the \emph{map} stage summarizes individual documents using the fixed instruction: \textit{"Summarize the given documents concisely, focusing on the key points and main ideas."}

The resulting summaries are concatenated into a single prompt. In the subsequent \emph{reduce} stage, the standard dataset templates (Figure~\ref{fig:loft_rag_prompt} or \ref{fig:loft_icl_prompt}) are used, with the concatenated summaries substituting the raw documents in the \texttt{\{context\}} slot.

\subsection{LongBench}
\label{sec:appendix-longbench}
For LongBench~\citep{bai-etal-2024-longbench}, we strictly adhere to the official task instructions and question templates outlined in the original paper's Appendix B. 
Dataset statistics (e.g., number of examples, context lengths, and task distributions) are reported in Table~1 of the original LongBench paper~\citep{bai-etal-2024-longbench}.

\subsection{Synthetic Dataset}
To benchmark TTFT and end-to-end latency (Figure~\ref{fig:ttft}) under controlled context length, we construct a small synthetic dataset with fixed formatting and token budgets. Each instance contains $N{=}64$ documents; exactly one \emph{gold} document includes a ``secret code'' string, while the remaining documents are distractors. We enforce an exact document length of 2048 tokens via padding/truncation. The query asks the model to output the secret code verbatim. We include a warmup sample to eliminate one-time initialization effects and stabilize latency measurements. 

\section{Normalization of Retrieval and Reranker Scores}
\label{sec:norm_scores}

\paragraph{Motivation.}
\textsc{Pced} uses retrieval and reranker scores as a document-level prior (Eq.~\ref{eq:contrastive-retrieval-bias-wrapped}),
where the prior enters as $\log r_k$. We therefore map all relevance signals to a common range $r_k \in [0,1)$ (and
clip away from 0 to avoid $\log 0$).

\paragraph{Retrieval scores (BGE-M3).}
Let $s_k$ denote the raw retrieval score produced by \texttt{bge-m3} for expert $k$ under a given scoring mode.
Different modes have different score ranges, so we normalize as follows:

\textbf{Dense / ColBERT.}
For the \texttt{dense} and \texttt{colbert} modes, similarity scores are bounded in $[-1,1]$.
We apply an affine rescaling:
\begin{equation}
\label{eq:norm_dense}
r_k^{\text{ret}} = \mathrm{clip}\!\left(\frac{s_k + 1}{2},\, 0,\, 1-\epsilon \right),
\end{equation}
which maps $[-1,1]\mapsto[0,1]$, followed by clipping to $[0,1-\epsilon)$.

\textbf{Sparse.}
For the \texttt{sparse} mode, scores are nonnegative and unbounded. Following standard practice for normalizing
unbounded similarity/distance values into $[0,1]$ (e.g., arctan-based normalization used in hybrid reranking),
we apply a saturating arctan transform:
\begin{equation}
\label{eq:norm_sparse}
r_k^{\text{ret}} = \mathrm{clip}\!\left(\frac{2}{\pi}\arctan(\max(s_k,0)),\, 0,\, 1-\epsilon \right).
\end{equation}
This preserves monotonicity while smoothly compressing large sparse scores.

\paragraph{Reranker scores (BGE reranker).}
We use \texttt{BAAI/bge-reranker-v2-m3} via \texttt{FlagReranker}. With \texttt{normalize=True}, the reranker
applies a sigmoid to map raw logits to $[0,1]$:
\begin{equation}
\label{eq:norm_rer}
r_k^{\text{rer}} = \sigma(z_k) = \frac{1}{1+\exp(-z_k)}.
\end{equation}
As above, we clip to $[0,1-\epsilon)$ before using the values in $\log r_k$.

\paragraph{Score fusion.}
After normalization, we combine retrieval and reranker signals into a single relevance score 
using the harmonic mean:
\begin{equation}
\label{eq:hmean_appendix}
r_k = \frac{2\, r_k^{\text{ret}}\, r_k^{\text{rer}}}{r_k^{\text{ret}} + r_k^{\text{rer}} + \epsilon}.
\end{equation}

\noindent In all experiments we set $\epsilon=10^{-8}$.
\section{Ablations}
\label{sec:ablations}

In this section, we provide a detailed analysis of the hyperparameters governing \textsc{Pced}. Unless otherwise stated, all ablations are performed using the \textbf{\textsc{Pced}-Dense} variant on the \textsc{HotpotQA} and \textsc{Natural Questions (NQ)} datasets, using both \textsc{Llama-3.1-8B-Instruct} and \textsc{Mistral-Nemo-13B-Instruct}.

\subsection{Impact of Contrastive Strength ($\beta$)}
\label{sec:beta_ablation}

The contrastive strength parameter $\beta$ determines how aggressively the expert distribution ($s_k$) is sharpened against the amateur prior ($s_0$). We compare our default dynamic $\beta$ strategy (derived from AdaCAD) against fixed values $\beta \in \{0.25, 0.5, 0.75, 1.0\}$. Additionally, we evaluate the setting $\beta=0$, which effectively removes the contrastive component and relies solely on the retrieval prior and raw expert logits.

Results are presented in Table~\ref{tab:ablation_beta}. We observe three key trends:

\begin{enumerate}
    \item \textbf{Necessity of Contrastive Decoding ($\beta > 0$):} Setting $\beta=0$ generally degrades performance compared to the best contrastive settings, confirming that subtracting the amateur logit helps isolate the specific knowledge provided by the retrieved document.
    \item \textbf{Instability of Fixed $\beta$:} While specific fixed values can achieve high scores on individual tasks (e.g., $\beta=0.25$ on Llama-NQ or $\beta=0.75$ on Llama-HotpotQA), they are inconsistent. A value that works well for one dataset may fail on another (e.g., $\beta=0.75$ drops significantly on Mistral-HotpotQA compared to lower values).
    \item \textbf{Robustness of Dynamic $\beta$:} The dynamic strategy consistently delivers competitive performance across all models and datasets without requiring per-task tuning. We therefore select \textbf{Dynamic} as the default to ensure stability across diverse retrieval scenarios.
\end{enumerate}

\begin{table}[h]
    \centering
    \caption{\textbf{Ablation of Contrastive Strength ($\beta$).} We compare fixed $\beta$ values against our Dynamic strategy. The column $\beta=0$ represents standard decoding without the contrastive penalty (only retrieval prior). \textbf{Bold} denotes the best result.}
    \label{tab:ablation_beta}
    \resizebox{\columnwidth}{!}{
    \begin{tabular}{ll c cccc >{\columncolor{pedbg}}c}
        \toprule
        & & \textbf{No CD} & \multicolumn{4}{c}{\textbf{Fixed} $\beta$} & \textbf{Ours} \\
        \cmidrule(lr){3-3} \cmidrule(lr){4-7} \cmidrule(lr){8-8}
        \textbf{Model} & \textbf{Dataset} & $\beta=0$ & 0.25 & 0.50 & 0.75 & 1.0 & \textbf{Dynamic} \\
        \midrule
        \multirow{2}{*}{\textsc{Llama-8B}} 
          & \textsc{HotpotQA} & 53 & 65 & 61 & \textbf{67} & 59 & 64 \\
          & \textsc{NQ}       & 70 & \textbf{88} & 65 & 84 & 62 & 85 \\
        \midrule
        \multirow{2}{*}{\textsc{Mistral-13B}} 
          & \textsc{HotpotQA} & 65 & 62 & 62 & 58 & 54 & \textbf{66} \\
          & \textsc{NQ}       & 80 & \textbf{83} & 82 & 78 & 80 & 81 \\
        \bottomrule
    \end{tabular}
    }
\end{table}

\subsection{Sensitivity to Retrieval Prior ($\gamma$)}
\label{sec:gamma_ablation}

The parameter $\gamma$ controls the influence of the retrieval/reranker scores on expert selection via the term $\gamma \log r_k$. We perform a sweep over $\gamma \in \{0.5, 1.0, 1.5, 2.0, 3.0, 4.0\}$ and compare these with our chosen default $\gamma=2.5$.

Results are shown in Table~\ref{tab:ablation_gamma}. We observe the following:

\begin{itemize}
    \item \textbf{Under-weighting ($\gamma < 1.5$):} Lower values often degrade performance (e.g., Llama-NQ drops significantly to 75 at $\gamma=0.5$). This confirms that expert selection cannot rely on internal perplexity alone; strong external relevance signals are necessary to suppress distractors.
    \item \textbf{Over-weighting ($\gamma \ge 4.0$):} High values yield inconsistent results. While Llama-NQ peaks at $\gamma=4.0$ (87), Llama-HotpotQA degrades compared to lower values (64 vs 66). Excessive gating forces the model to rigidly follow the retriever's ranking, potentially overriding valid reasoning from lower-ranked experts on complex queries.
    \item \textbf{$\gamma=2.5$:} The range $\gamma \in [2.0, 3.0]$ represents a stable "sweet spot" across both models and datasets. We select $\gamma=2.5$ as the default because it offers the best trade-off: it maximizes performance on difficult tasks like NQ (matching the high scores of $\gamma=2.0-3.0$) while avoiding the instability seen at the extremes.
\end{itemize}

\begin{table}[h]
    \centering
    \caption{\textbf{Sensitivity sweep for Retrieval Prior weight ($\gamma$).} We use Dynamic $\beta$ for all runs. The main paper uses $\gamma=2.5$.}
    \label{tab:ablation_gamma}
    \setlength{\tabcolsep}{3.5pt}
    \resizebox{\columnwidth}{!}{
    \begin{tabular}{ll cccccc c}
        \toprule
         & & \multicolumn{6}{c}{\textbf{Gamma} ($\gamma$)} & \textbf{Default} \\
        \cmidrule(lr){3-8} \cmidrule(lr){9-9}
        \textbf{Model} & \textbf{Dataset} & 0.5 & 1.0 & 1.5 & 2.0 & 3.0 & 4.0 & \textbf{2.5} \\
        \midrule
        \multirow{2}{*}{\textsc{Llama-8B}} 
          & \textsc{HotpotQA} & 65 & \textbf{66} & \textbf{66} & 65 & 63 & 64 & 64 \\
          & \textsc{NQ}       & 75 & 84 & 86 & 85 & 85 & \textbf{87} & 85 \\
        \midrule
        \multirow{2}{*}{\textsc{Mistral-13B}} 
          & \textsc{HotpotQA}       & 64 & 65 & 66 & 65 & \textbf{67} & 66 & 66 \\
          & \textsc{NQ} & 78 & 79 & 80 & \textbf{81} & \textbf{81} & \textbf{81} & 81 \\
        \bottomrule
    \end{tabular}
    }
\end{table}
\begin{figure*}[t]
    \centering
    \includegraphics[width=\linewidth]{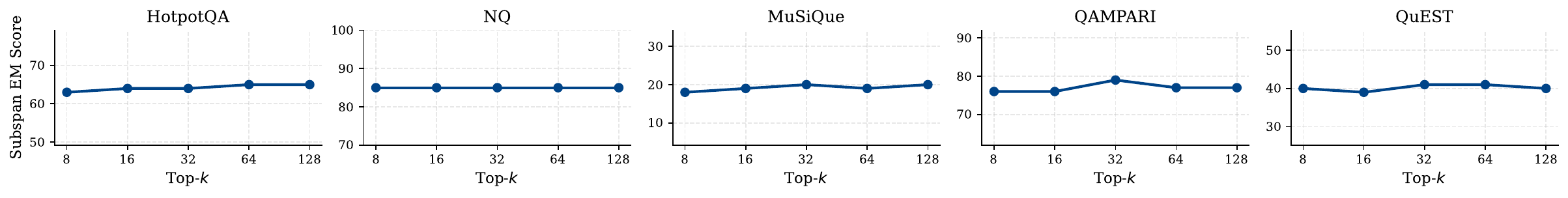}
    \caption{\textbf{Performance Stability across Top-$k$.} \textsc{Pced} maintains consistent accuracy from $k=8$ to $128$, confirming that the retrieval prior effectively suppresses noise from additional distractors.}
    \label{fig:topk}
\end{figure*}
\subsection{Contrastive Signal vs. Retrieval Score Only}
\label{sec:retrieval_only_ablation}

Finally, we isolate the contribution of the two core components of Equation~\ref{eq:contrastive-retrieval-bias-wrapped}: the contrastive signal ($\beta$) and the retrieval prior ($\gamma$). 

Table~\ref{tab:ablation_components} compares the full \textsc{Pced} method against two ablations:
\begin{enumerate}
    \item \textbf{Only Retrieval Scores ($\beta=0$):} Expert logits are boosted by retrieval scores but not calibrated against the amateur.
    \item \textbf{Only Contrastive ($\gamma=0$):} Expert logits are calibrated via contrastive decoding, but all experts are treated as equally likely (flat prior), ignoring retrieval ranking.
\end{enumerate}

The results reveal two distinct findings:
\begin{itemize}
    \item \textbf{Retrieval Prior is Foundational ($\gamma > 0$):} The "Only Contrastive" setting fails catastrophically across all benchmarks (e.g., Llama NQ drops to 52). This confirms that without the external guidance of the retriever to gate irrelevant experts, the model is overwhelmed by noise from distractors.
    \item \textbf{Contrastive Signal is an Amplifier ($\beta > 0$):} The impact of the contrastive term is model-dependent. For \textsc{Llama-3.1}, it is critical: removing it ("Only Retrieval") causes a massive drop (e.g., NQ falls from 85 to 70), suggesting that Llama requires the amateur subtraction to suppress its own priors and hallucinations. Conversely, \textsc{Mistral} is more robust, achieving strong performance with retrieval scores alone, though the full \textsc{Pced} framework still secures the highest absolute scores in all cases.
\end{itemize}

\begin{table}[h]
    \centering
    \caption{\textbf{Component Analysis.} We disentangle the benefits of the Contrastive Decoding signal versus the Retrieval Prior.}
    \label{tab:ablation_components}
    \resizebox{\columnwidth}{!}{
    \begin{tabular}{ll ccc}
        \toprule
        & & \textbf{Only Contrastive} & \textbf{Only Retrieval} & \textbf{Full} \\
        & & (No Prior, $\gamma=0$) & (No CD, $\beta=0$) & \textbf{PCED} \\
        \midrule
        \multirow{2}{*}{\textsc{Llama-8B}} 
          & \textsc{HotpotQA} & 46 & 53 & \textbf{64} \\
          & \textsc{NQ}       & 52 & 70 & \textbf{85} \\
        \midrule
        \multirow{2}{*}{\textsc{Mistral-13B}} 
          & \textsc{HotpotQA} & 57 & 65 & \textbf{66} \\
          & \textsc{NQ}       & 71 & 80 & \textbf{81} \\
        \bottomrule
    \end{tabular}
    }
\end{table}
\subsection{Ablation of Expert Aggregation Rule}
\label{sec:agg_ablation}

\textsc{Pced} aggregates experts via a token-wise \textbf{Max} operation. We compare this against two probability-space alternatives: \textbf{Mixture-of-Experts} (MoE, weighted sum) and \textbf{Product-of-Experts} (PoE, weighted product), where weights are derived from retrieval scores.

Table~\ref{tab:ablation_aggregation} shows that \textbf{Max} aggregation is critical for multi-hop reasoning (\textsc{HotpotQA}), outperforming MoE by 8 points (64 vs. 56). We hypothesize that Max enables sharper \emph{token-level expert switching}, allowing different documents to dominate different generation steps without their distributions needing to agree. Conversely, on single-document tasks like \textsc{NQ}, MoE performs slightly better (87 vs. 85), suggesting that soft averaging can be beneficial when evidence is concentrated in one expert and retrieval priors are accurate.

\begin{table}[h]
    \centering
    \caption{\textbf{Aggregation Rule Ablation.} Comparison of \textsc{Pced} (Max) vs. probabilistic aggregation (MoE, PoE).}
    \label{tab:ablation_aggregation}
    \setlength{\tabcolsep}{6pt}
    \begin{tabular}{lcc}
        \toprule
        \textbf{Aggregation} & \textsc{HotpotQA} & \textsc{NQ} \\
        \midrule
        \textbf{Max (Ours)} & \textbf{64} & 85 \\
        Mixture (MoE) & 56 & \textbf{87} \\
        Product (PoE) & 46 & 85 \\
        \bottomrule
    \end{tabular}
\end{table}

\subsection{Robustness to Candidate Pool Size ($k$)}
\label{sec:topk_ablation}
We evaluate the stability of \textsc{Pced} (Dense, Llama-3.1-8B) as we scale the number of retrieved experts from $k=8$ to $k=128$. Results are visualized in Figure~\ref{fig:topk}.

We observe two trends:
\begin{itemize}
    \item \textbf{Noise Tolerance:} Performance remains nearly constant across all datasets despite a $16\times$ increase in experts. For instance, \textsc{NQ} scores stay flat at $\sim$85, while \textsc{HotpotQA} fluctuates only marginally (63--65). This confirms that the retrieval prior ($\gamma \log r_k$) effectively gates low-relevance experts, preventing distractor accumulation.
    \item \textbf{Recall without Penalty:} While low $k$ is often sufficient, the lack of degradation at $k=128$ allows users to maximize recall for difficult queries without sacrificing generation quality.
\end{itemize}

\end{document}